\documentclass{article}

\usepackage{arxiv}

\usepackage[utf8]{inputenc} 
\usepackage[T1]{fontenc}    
\usepackage{hyperref}       
\usepackage{url}            
\usepackage{booktabs}       
\usepackage{amsfonts}       
\usepackage{amsmath}        
\usepackage{nicefrac}       
\usepackage{microtype}      
\usepackage{lipsum}		
\usepackage{graphicx}
\usepackage{natbib}
\usepackage{doi}

\usepackage{booktabs}
\usepackage{multirow}
\usepackage[table]{xcolor}
\usepackage{arydshln}
\usepackage{caption}
\usepackage{graphicx}
\usepackage{makecell}
\usepackage{subcaption}

\title{Cognitively-Inspired Tokens Overcome Egocentric Bias in Multimodal Models}

\author{ 
    Bridget Leonard\\ 
	Department of Psychology\\
	University of Washington\\
	Seattle, WA 98195 \\
	\texttt{bll313@uw.edu} \\
	\And
    Scott O. Murray\\ 
	Department of Psychology\\
	University of Washington\\
	Seattle, WA 98195 \\
	\texttt{somurray@uw.edu} \\
}

\date{}

\renewcommand{\headeright}{}
\renewcommand{\undertitle}{}

\hypersetup{
pdftitle={A template for the arxiv style},
pdfsubject={q-bio.NC, q-bio.QM},
pdfauthor={David S.~Hippocampus, Elias D.~Striatum},
pdfkeywords={First keyword, Second keyword, More},
}

\begin{document}
\maketitle

\begin{abstract}
	Multimodal language models (MLMs) perform well on semantic vision–language tasks but fail at spatial reasoning that requires adopting another agent’s visual perspective. These errors reflect a persistent egocentric bias and raise questions about whether current models support allocentric reasoning. Inspired by human spatial cognition, we introduce \textbf{\textit{perspective tokens}}, specialized embeddings that encode orientation through either (1) embodied body-keypoint cues or (2) abstract representations supporting mental rotation. Integrating these tokens into LLaVA-1.5-13B yields performance on level-2 visual perspective-taking tasks. Across synthetic and naturalistic benchmarks (Isle Bricks V2, COCO, 3DSRBench), perspective tokens improve accuracy, with rotation-based tokens generalizing to non-human reference agents. Representational analyses reveal that fine-tuning enhances latent orientation sensitivity already present in the base model, suggesting that MLMs contain precursors of allocentric reasoning but lack appropriate internal structure. Overall, embedding cognitively grounded spatial structure directly into token space provides a lightweight, model-agnostic mechanism for perspective-taking and more human-like spatial reasoning.
\end{abstract}


\section*{Introduction}

Spatial reasoning is a core process in human cognition. Engaging spatial reasoning often involves making perspective-based decisions, such as judging how others perceive the world or how our own perception changes as we move through space. A critical component of such reasoning is visual perspective-taking (VPT) — the ability to simulate a visual perspective other than one’s own — which plays a central role in social interaction and spatial navigation. If multimodal language models (MLMs) are to support perspective-based spatial reasoning, they must likewise be able to represent and operate over non-egocentric viewpoints. However, despite achieving near-human, and often superhuman, performance on many vision-language tasks, MLMs consistently fail at perspective-based reasoning, instead defaulting to an egocentric perspective \citep{zhang2025sphereunveilingspatialblind,zhang2025visionlanguagemodelsrepresentspace,leonard2024failuresperspectivetakingmultimodalai,góral2024seeingeyesevaluatingvisual,góral2025recognitionevaluatingvisualperspective}. 

This failure is particularly striking given that an egocentric bias is a well-established feature of human cognition, but one that humans can systematically overcome through development and experience. Understanding how and when humans overcome this bias offers a unique opportunity to investigate MLM capabilities through the lens of cognitive development. In the human developmental literature, VPT is divided into two levels that emerge sequentially. While level 1 VPT, understanding what another person can or cannot see, develops around two years of age \citep{moll2006level}, understanding how a scene appears from another's viewpoint, or level 2 VPT, begins to develop at age four \citep{newcombe1992children} and continues throughout middle childhood \citep{surtees2012egocentrism} and even into adolescence and young adulthood \citep{dumontheil2010taking}. This protracted developmental trajectory suggests that level 2 VPT depends on cognitive processes that are qualitatively distinct from those supporting level 1 VPT, and that these processes require extended experience to mature. If humans, with their rich embodied experience and social learning, need years to develop this skill, it is perhaps unsurprising that MLMs trained primarily on static image-text pairs struggle with perspective-taking. Understanding the cognitive mechanisms humans employ to overcome egocentric bias may provide insight into what is fundamentally missing in current MLMs, and suggest pathways for developing these capabilities.

Researchers have explored several approaches to enable VPT in MLMs. One proposed solution has been to augment MLMs with external computer vision systems for spatial reasoning. For example, \cite{lee2025perspectiveawarereasoningvisionlanguagemodels} developed Abstract Perspective Change (APC) in which vision foundation models (e.g., depth, pose, orientation estimation) parse perspective-based prompts to guide reasoning in language models. While effective, such multi-model approaches are computationally expensive and, critically, do not address whether spatial reasoning can emerge within the internal representations of MLMs themselves. An alternative line of work suggests that visual grounding can be strengthened through explicit token design. LLaVA Aurora \citep{bigverdi2024perceptiontokensenhancevisual} introduces perception tokens that encode intermediate visual representations, such as depth maps and bounding box coordinates, directly into the language model’s token space. By expanding the tokenizer and fine-tuning the multimodal projector and language embeddings, this approach allows models to reason over structured visual information internally, improving performance on perception-heavy tasks without relying on external tools. These findings raise a compelling question: if perception tokens can enhance low-level visual understanding, could similarly structured tokens help MLMs acquire the higher-order spatial reasoning skills required for perspective-taking?

In this study, we leveraged insights from human visual perspective-taking (VPT) to test whether perspective-based reasoning can emerge within the internal representations of a multimodal language model. Specifically, we focused on LLaVA 1.5 13B \citep{liu2023improvedllava}, a relatively small, transformer-based, open-source MLM. In the human cognitive literature, level 2 VPT relies on cognitive processes that are qualitatively distinct from those involved in level 1 VPT. Behaviorally, response times in level 2 VPT tasks increase systematically with the degree of misalignment between the observer and the reference frame, whereas the degree of alignment has little effect on level 1 judgments \citep{surtees2013similarities}. Neuroimaging studies further show that level 2 VPT engages a distributed network including regions associated with theory of mind, internal body representations, and mental rotation \citep{GUNIA2021113247}. Together, these findings suggest that humans overcome egocentric bias in level 2 VPT by temporarily embodying another perspective and transforming spatial representations accordingly. We used these mechanistic insights as a guiding framework to probe whether analogous internal operations could be induced in a transformer-based MLM, rather than relying on external spatial reasoning modules.

Inspired by VPT as an embodied process \citep{kessler2010embodied}, we developed two complementary approaches grounded in distinct cognitive mechanisms observed in human perspective-taking. The first approach focused on perspective embodiment, using tokens derived from body-related information to support reasoning about whether a reference was in an aligned (same perspective as viewer) or unaligned (opposite perspective as viewer) state. This approach allowed us to examine whether explicit sensitivity to perspective alignment could support allocentric reasoning within the model. Observations that some degree of alignment sensitivity was present even in the base model motivated a second approach that treats VPT as a form of mental rotation. In this framework, humans construct abstract scene representations that can be transformed to reconcile differing viewpoints \citep{surtees2013use}. Consequently, our second approach incorporated spatial information about both reference and query objects to enable direct transformation of scene representations based on the reference’s orientation. Together, these approaches provide a mechanistically grounded test of whether human-inspired cognitive operations can be instantiated within MLM representations, offering a principled framework for comparing artificial and human perspective-taking.

\section*{Results}

\begin{figure}[ht]
\centering
\includegraphics[width=0.5\linewidth]{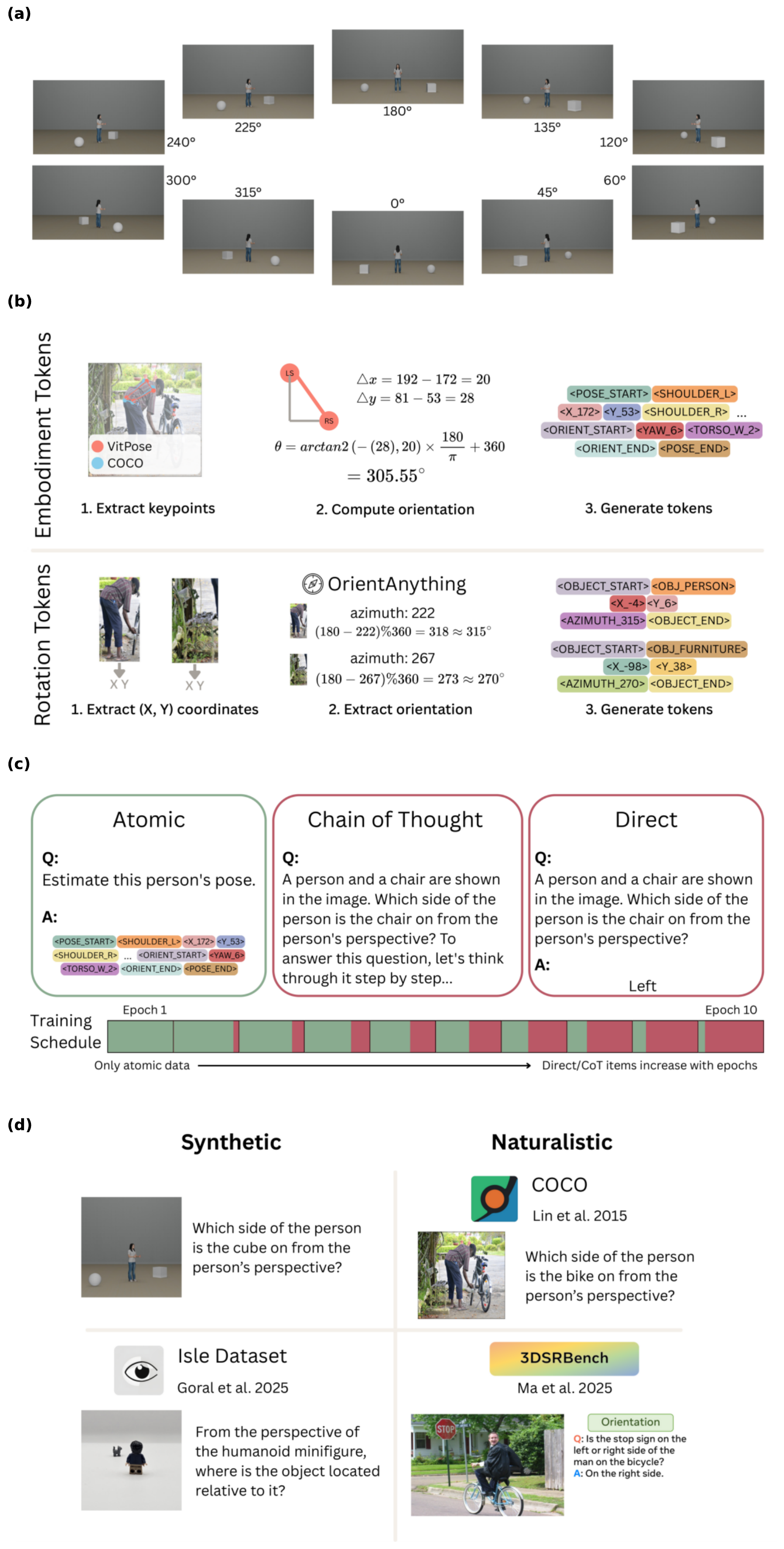}
\caption{Overview of evaluation and training data. \textbf{a}, Stimuli from the perspective-taking benchmark systematically vary the reference avatar's orientation across degrees of alignment (0°–360°). \textbf{b}, Token construction methods: embodiment tokens encode body keypoint coordinates and calculated yaw orientation (above); rotation tokens encode object bounding box coordinates and azimuth labels from OrientAnything (below). \textbf{c}, Curriculum learning structure showing progression from atomic token generation to chain-of-thought reasoning to direct answer formats. The training schedule increases the proportion of chain-of-thought and direct items by 10\% each epoch. \textbf{d}, Representative items from each evaluation benchmark (Perspective-Taking, Isle Bricks V2, COCO validation, 3DSRBench).}
\label{fig1}
\end{figure}

In our analysis, we primarily evaluated our models on a perspective-taking benchmark that systematically varies the reference's alignment with the viewer's perspective (Fig. \ref{fig1}a). Alignment refers to the degree of orientation similarity: references at 0° are fully aligned (identical spatial judgments), while references at 180° are maximally unaligned (reversed judgments). Both our base model for perspective tokens, LLaVA 1.5 13B, and state-of-the-art models, such as GPT-4o, characteristically fail at unaligned angles (\citealt{leonard2024failuresperspectivetakingmultimodalai}; Fig. \ref{fig2}a). We further evaluated our models on a variety of benchmarks containing level 2 VPT tasks, including Isle Bricks V2 \citep{góral2025recognitionevaluatingvisualperspective}, 3DSRBench \citep{ma20253dsrbenchcomprehensive3dspatial}, and on held-out COCO images. The perspective-taking and Isle Bricks benchmarks represent the most controlled evaluations in our study, with reference orientation systematically manipulated in synthetic stimuli. To assess generalization to naturalistic settings, we additionally evaluated performance on 3DSRBench and held-out COCO images, which contain real-world scenes with greater visual complexity and variability (Fig. \ref{fig1}d).

As expected, on all three additional benchmarks, LLaVA 1.5 13B succeeded on aligned items while failing at unaligned angles (Fig. \ref{fig2}d). Interestingly, all of the unaligned Isle Bricks V2 items that LLaVA got correct (27\%) were bird's-eye view images.

\subsection*{Embodiment tokens reduce egocentric bias in perspective-taking tasks}

\begin{figure}[ht]
\centering
\includegraphics[width=0.45\linewidth]{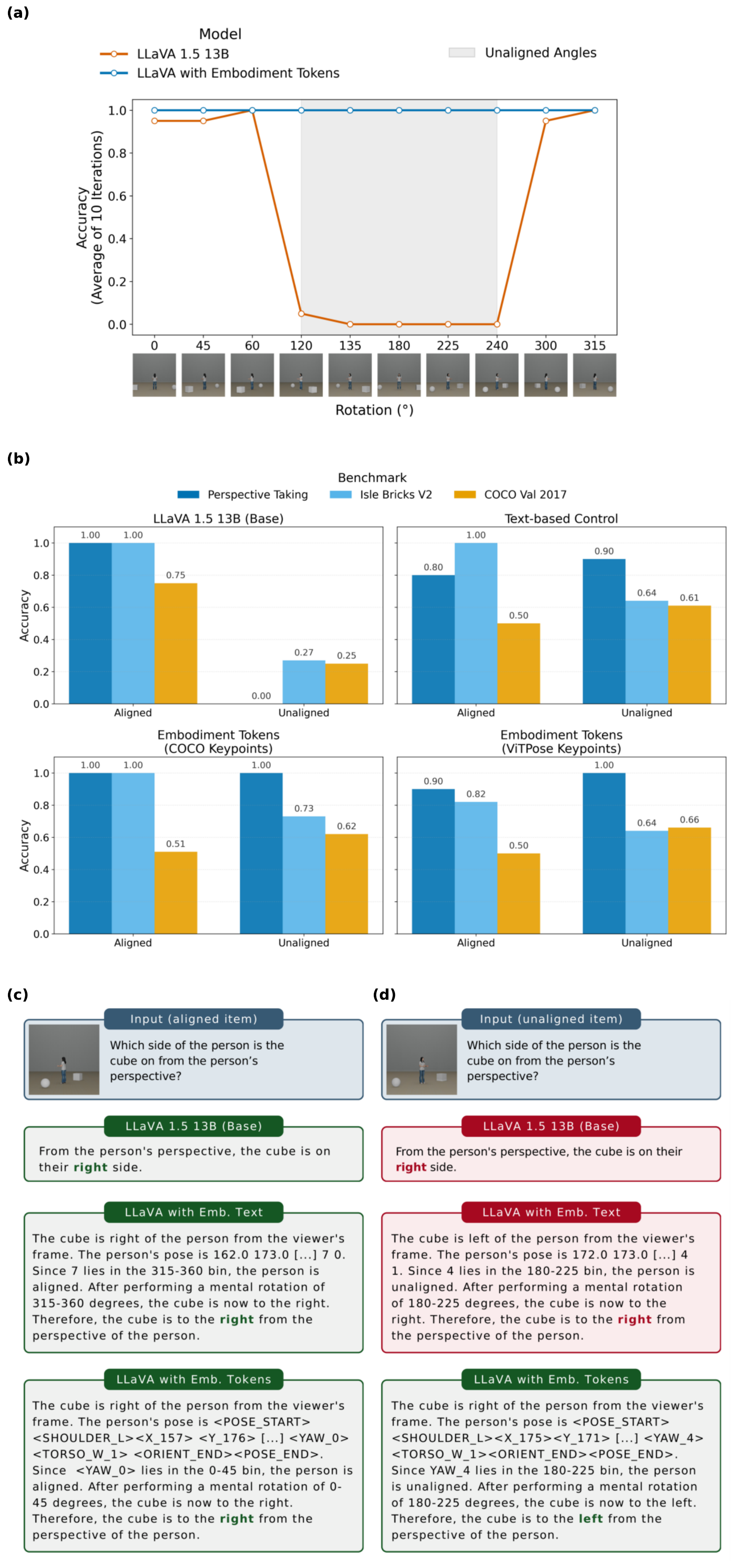}
\caption{LLaVA with embodiment tokens succeeds at unaligned perspective-taking tasks. \textbf{a}, LLaVA 1.5 13B fails at unaligned angles on the perspective-taking benchmark (left) while LLaVA with embodiment tokens succeeds (right). \textbf{b}, Performance across benchmarks: LLaVA with embodiment tokens substantially outperforms the base model on unaligned items from the perspective-taking benchmark, Isle Bricks V2, and COCO validation sets. The text-based control (embodiment information presented in natural language) improves over baseline but falls short of token-based performance. \textbf{c-d}, Response comparisons for aligned \textbf{(c)} and unaligned \textbf{(d)} queries showing base model (direct prompt), text-based model (chain-of-thought), and token-based model (chain-of-thought).}
\label{fig2}
\end{figure}

We first evaluated LLaVA with embodiment tokens on our perspective-taking benchmark to assess performance at unaligned angles. Embodiment tokens encode the reference's body keypoint locations (shoulders, hips) as specialized tokens, allowing the model to explicitly represent the reference's orientation and torso alignment relative to the viewer. These tokens are structured sequentially: keypoint coordinates are encoded first, followed by a yaw token that the model learns to predict from the keypoint configuration, establishing an explicit link between body pose and orientation (Fig. \ref{fig1}b). Reported improvements reflect the difference between base model performance and the average performance of the fine-tuned model across two prompting conditions: direct prompting (where the model responds immediately) and chain-of-thought prompting (CoT; where the model explicitly reasons through the perspective transformation), similar to the curriculum data conditions (Fig. \ref{fig1}c). High performance under chain-of-thought prompting indicates that the model has learned to reason with embodiment tokens, while high performance under direct prompting suggests internalization of the perspective-taking process without requiring explicit reasoning steps. Embodiment tokens significantly improved performance, with LLaVA reaching full accuracy on unaligned items (+100\% improvement) and 95\% overall (+48\%).  Figure \ref{fig2}c illustrates this improvement, contrasting the base model's egocentric response with the embodiment token model's correct allocentric reasoning (abbreviated response shown). This reduction in egocentric bias and ability to adopt an allocentric perspective generalized to other evaluations, including Isle Bricks (+50\% on unaligned; +33.3\% total) and held-out COCO images (+72.2\% on unaligned; +47.9\% total) (Fig. \ref{fig2}). However, performance on 3DSRBench showed little improvement likely due to the prevalence of non-human references (e.g., animals) that embodiment tokens, grounded in human body keypoints, cannot adequately represent.

To isolate the contribution of specialized token embeddings, we created a text-based control condition containing the same information as embodiment tokens but in natural language form, leaving the model's tokenizer unchanged during fine-tuning. This text-based model outperformed the base model on all benchmarks (Fig. \ref{fig2}d) but fell short of achieving token-based performance on perspective-taking and Isle Bricks, which included the most complexities of angles of alignment (+15\% on perspective-taking; +22.7\% on Isle Bricks). This dissociation demonstrates that while embodiment information is beneficial in any form, a dedicated embedding space for spatial information, provided by specialized tokens, is necessary for the model to fully overcome egocentric bias. However, the text-based model outperformed the token model on both COCO validation and 3DSRBench (+53.1 \% COCO val; +22.2\% 3DSRBench), possibly reflecting that text provides increased flexibility in more complex scenes. Comparisons between the base, text, and different keypoint token models are detailed in Table 1.

\subsection*{Fine-tuning amplifies alignment sensitivity present in base MLMs}

\begin{figure}[ht]
\centering
\includegraphics[width=0.5\linewidth]{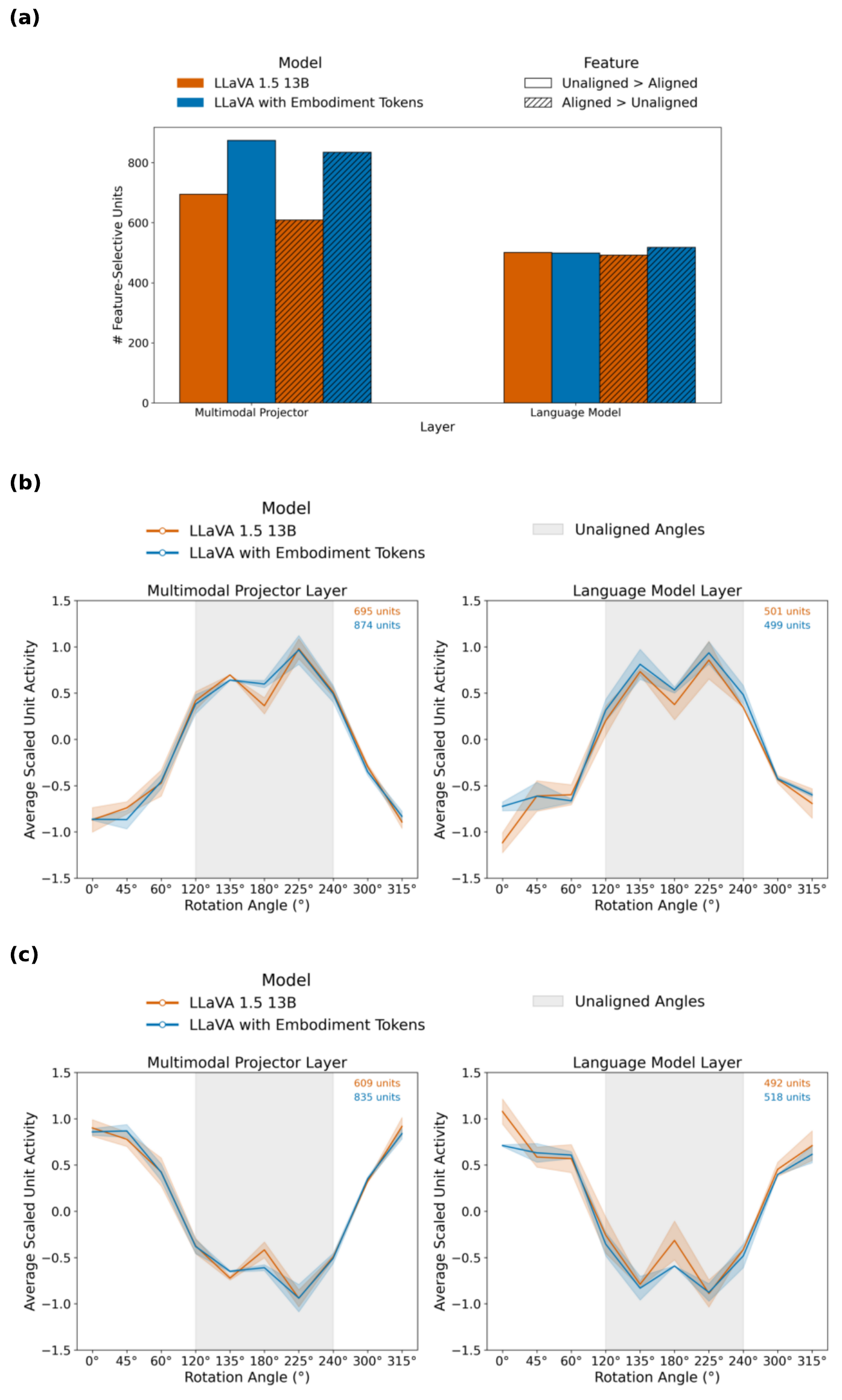}
\caption{Intermediate activations show alignment sensitivity in both embodiment token and base LLaVA. \textbf{a}, Number of feature selective units in base LLaVA 1.5 13B versus embodiment token model, measured at the multimodal projector layer (left) and language model layer (right). \textbf{b-c}, Average activation profiles of alignment-selective units as a function of reference angle, for units preferring unaligned over aligned conditions \textbf{(b)} and units preferring aligned over unaligned conditions \textbf{(c)}. The approximately mirror-symmetric activation patterns reflect the use of standardized (z-scored) unit activations, such that preference for one condition implies reduced activation for the opposite condition.}
\label{fig3}
\end{figure}

To investigate how internal representations change with fine-tuning, we examined hidden unit activations in response to controlled stimulus features from the perspective-taking benchmark (avatar alignment, angle of difference, cube direction). We specifically found that both the base LLaVA model and the LLaVA model with embodiment tokens contained a substantial number of feature selective units for both \textit{unaligned $>$ aligned} and \textit{aligned $>$ unaligned} contrasts, with the embodiment token model typically showing a greater number of such units (Fig. \ref{fig3}a).

Although these units were identified using a binary aligned vs. unaligned comparison, their activation profiles were not dichotomous. Units from either contrast showed smoothly varying responses across angular differences, consistent with graded orientation tuning rather than categorical aligned/unaligned coding (Fig. \ref{fig3}b, Fig. \ref{fig3}c). Notably, activation patterns for the two feature selection types appeared roughly inverse, a pattern that reflects the use of standardized neural activity (z-scored per unit), where selecting units with positive preference for one condition necessarily implies a negative preference for the other. This tuning pattern emerged in both multimodal- and language-layer feature selective units, and critically, in both base and fine-tuned models, though the base model had fewer feature selective units identified.

The presence of alignment sensitivity in the base model suggested two opportunities for our next approach. First, if the base model already encodes orientation information, a lighter-weight method that directly labels reference orientation might suffice, bypassing keypoint extraction overhead. Second, by encoding multiple objects with their spatial orientations, we could create an abstract scene representation analogous to the mental rotation process humans employ during level 2 VPT.

\subsection*{Direct spatial encoding supports flexible perspective-taking}

\begin{figure}[ht]
\centering
\includegraphics[width=0.35\linewidth]{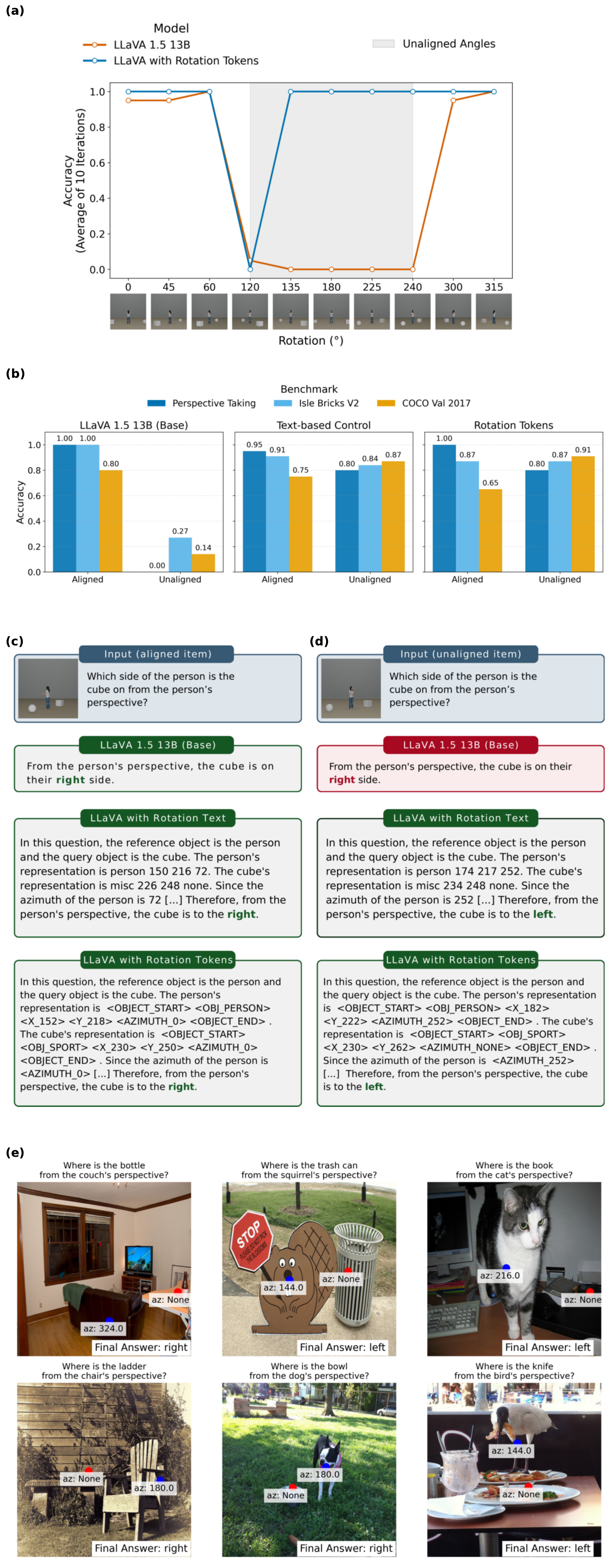}
\caption{Similar to embodiment tokens, LLaVA with rotation tokens succeeded at unaligned perspective-taking tasks. \textbf{a}, Unlike LLaVA 1.5 13B (left), LLaVA with rotation tokens succeeded on unaligned items on the perspective-taking benchmark (right). \textbf{b}, Performance across benchmarks: LLaVA with rotation tokens substantially outperforms the base model on unaligned items from the perspective-taking benchmark, Isle Bricks V2, and COCO validation sets. The text-based control (embodiment information in natural language) improves over baseline but falls short of token-based performance. \textbf{c-d}, Response comparisons for aligned \textbf{(c)} and unaligned \textbf{(d)} queries showing base model (direct prompt), text-based model (chain-of-thought), and token-based model (chain-of-thought). \textbf{e,} Examples of LLaVA with rotation tokens generalizing to non-human references like furniture and animals from 3DSRBench.}
\label{fig4}
\end{figure}

Rotation tokens, which encode abstract scene information with explicit orientation labels (Fig. \ref{fig1}b), successfully corrected LLaVA's performance on our perspective-taking benchmark (+30\% unaligned; +40\% total; Fig. \ref{fig4}a). This is particularly notable given that rotation tokens contain no information about how to categorize object orientation as aligned or unaligned, unlike embodiment tokens. Chain-of-thought analysis revealed that the single failure case (120°) resulted from misattribution to a nearby aligned angle, as this example falls close to the reversal point between aligned and unaligned states.

Similar to embodiment tokens, LLaVA with rotation tokens improved performance on the Isle Bricks benchmark (+59.1\% on unaligned; +34.8\% total) and held-out COCO images (+77.1\% on unaligned; +56.7\% total). Critically, since rotation tokens generalize beyond human body configurations, performance also improved on 3DSRBench (+21.1\%), with the model successfully adopting the perspectives of animals and chair-like furniture (Fig. \ref{fig4}d). 

Compared to rotation tokens, our text-based control performed better than the base model at all evaluations (+32.5\% Perspective-Taking; +34.8\% Isle Bricks; +55.5\% COCO val; +22.2\% 3DSRBench) but failed to reach token-based performance on the perspective-taking benchmark and held-out COCO images. Again, this demonstrates that while text-based spatial information is useful for the model in VPT, a specific embedding space allows MLM VPT abilities to reach the precision observed in human cognition. Comparisons between the base, text, and rotation token models are detailed in Table 1.

Our results suggest that perspective-taking does not simply emerge from larger-scale training or improved image encoders, but instead requires a specific representational substrate that supports viewpoint transformation. By embedding orientation-relevant structure directly into the token space, we provide the model with the equivalent of the embodied and rotational representations that humans rely on during level 2 VPT. This reveals that MLMs already contain latent cues relevant to allocentric reasoning, but these cues must be organized into a coherent representational format before perspective-taking can reliably emerge.




\begin{table}[htbp]
\centering
\renewcommand{\arraystretch}{1.2}
\setlength\dashlinedash{0.5pt}
\setlength\dashlinegap{1pt}
\setlength\arrayrulewidth{0.3pt}

\begin{tabular}{ll  c  ccc  ccc  ccc}
\toprule
\textbf{Evaluation} & & \textbf{LLaVA} & \multicolumn{3}{c}{\textbf{Emb. Text}} & \multicolumn{3}{c}{\makecell{\textbf{Emb. Tokens}\\\textbf{(COCO)}}} & \multicolumn{3}{c}{\makecell{\textbf{Emb. Tokens}\\\textbf{(ViTPose)}}} \\
\cmidrule(lr){3-3} \cmidrule(lr){4-6} \cmidrule(lr){7-9} \cmidrule(lr){10-12}
& & 1.5-13B & Direct & CoT & \textbf{Avg} & Direct & CoT & \textbf{Avg} & Direct & CoT & \textbf{Avg} \\
\midrule

\multirow{3}{*}{\makecell[l]{Perspective\\Taking}} 
  & Align.          & 1.00 & 0.80 & 0.80 & 0.80 & 1.00 & 0.90 & 0.95 & 0.90 & 1.00 & 0.95 \\
  & Unalign.        & 0.00 & 0.90 & 0.80 & 0.85 & 1.00 & 0.90 & 0.95 & 1.00 & 1.00 & 1.00 \\
  & \textbf{Total}  & 0.50 & 0.85 & 0.80 & 0.83 & 1.00 & 0.90 & 0.95 & 0.95 & 1.00 & \textbf{0.98} \\
  
\midrule

\multirow{3}{*}{\makecell[l]{Isle Bricks\\V2}}
  & Align.           & 1.00 & 1.00 & 0.82 & 0.91 & 1.00 & 1.00 & 1.00 & 0.82 & 0.82 & 0.82 \\
  & Unalign.         & 0.27 & 0.64 & 0.68 & 0.66 & 0.73 & 0.82 & 0.78 & 0.64 & 0.68 & 0.66 \\
  & \textbf{Total}   & 0.52 & 0.76 & 0.73 & 0.75 & 0.82 & 0.88 & \textbf{0.85} & 0.70 & 0.73 & 0.72 \\

\midrule

\multirow{3}{*}{\makecell[l]{COCO 2017\\Validation}} 
  & Align.             & 0.83 & 0.75 & 0.67 & 0.71 & 0.58 & 0.58 & 0.58 & 0.58 & 0.58 & 0.58 \\
  & Unalign.           & 0.14 & 0.83 & 0.94 & 0.89 & 0.81 & 0.92 & 0.86 & 0.86 & 0.83 & 0.84 \\
  & \textbf{Total}     & 0.31 & 0.81 & 0.88 & \textbf{0.84} & 0.75 & 0.83 & 0.79 & 0.79 & 0.77 & 0.78 \\

\midrule

3DSRBench & \textbf{Total} & 0.40 & 0.58 & 0.67 & \textbf{0.63} & 0.49 & 0.58 & 0.54 & 0.49 & 0.64 & 0.57 \\

\bottomrule
\end{tabular}

\vspace{0.5cm}

\begin{tabular}{ll  c  ccc  ccc}

\toprule
\textbf{Evaluation} & & \textbf{LLaVA} & \multicolumn{3}{c}{\textbf{Rotation Text}} & \multicolumn{3}{c}{\textbf{Rotation Tokens}} \\
\cmidrule(lr){3-3} \cmidrule(lr){4-6} \cmidrule(lr){7-9}
& & 1.5-13B & Direct & CoT & \textbf{Avg} & Direct & CoT & \textbf{Avg} \\
\midrule

\multirow{3}{*}{\makecell[l]{Perspective\\Taking}} 
  & Alig.              & 1.00 & 1.00 & 0.90 & 0.95 & 1.00 & 1.00 & 1.00 \\
  & Unalign.            & 0.00 & 0.90 & 0.70 & 0.80 & 0.80 & 0.80 & 0.80 \\
  & \textbf{Total}       & 0.50 & 0.95 & 0.80 & 0.88 & 0.90 & 0.90 & \textbf{0.90}\\
  
\midrule

\multirow{3}{*}{Isle Bricks V2} 
  & Align.              & 1.00 & 1.00 & 0.82 & 0.91 & 0.91 & 0.82 & 0.87 \\
  & Unalign.            & 0.27 & 0.86 & 0.82 & 0.84 & 0.82 & 0.91 & 0.87\\
  & \textbf{Total}     & 0.52 & 0.91 & 0.82 & \textbf{0.87} & 0.85 & 0.88 & \textbf{0.87} \\

\midrule

\multirow{3}{*}{\makecell[l]{COCO 2017\\Validation}} 
  & Align.              & 0.80 & 0.80 & 0.70 & 0.75 & 0.70 & 0.60 & 0.65 \\
  & Unalign.            & 0.14 & 0.89 & 0.86 & 0.87 & 0.89 & 0.94 & 0.91 \\
  & \textbf{Total}     & 0.29 & 0.87 & 0.82 & 0.84 & 0.84 & 0.87 & \textbf{0.86} \\

\midrule

3DSRBench &   \textbf{Total}     & 0.40 & 0.60 & 0.64 & \textbf{0.62} & 0.67 & 0.56 & \textbf{0.62} \\

\bottomrule
\end{tabular}

\vspace{0.5cm}
\caption{Model performance on evaluation benchmarks. Highest combined direct and CoT scores are bolded.}
\end{table}

\section*{Discussion}

The present study demonstrates that reducing egocentric bias in multimodal language models depends not simply on providing additional spatial information, but on organizing that information in ways that support perspective transformation. Drawing on concepts from cognitive psychology, we introduced spatially structured tokens inspired by prior work on perception tokens \citep{bigverdi2024perceptiontokensenhancevisual} to probe how human-relevant spatial representations might be instantiated with an MLM. Our first approach, embodiment tokens, captures aspects of visual perspective-taking as an embodied process by explicitly encoding body-based orientation cues, enabling the model to adopt allocentric perspectives when reasoning about human references. Analysis of the model’s internal representations revealed, however, that sensitivity to reference orientation was already present to some degree in the base model. This finding suggests that egocentric bias does not arise from an absence of relevant spatial signals, but from a failure to organize those signals into a form that supports flexible transformation. Building on this latent orientation sensitivity, our second approach introduced rotation tokens inspired by mental rotation processes, allowing spatial relations to be transformed directly rather than inferred through categorical alignment judgments. This more abstract representational strategy enabled perspective-taking to generalize beyond human bodies to non-human references, thereby overcoming a key limitation of embodiment-based representations.

These findings indicate that the persistent egocentric bias in current multimodal models reflects more than insufficient training scale or image-text supervision. They reveal a deeper representational gap: the absence of structured spatial encodings that support explicit transformation between viewpoints. Our results also differentiate perspective tokens from external-pipeline approaches that rely on dedicated vision modules for pose or depth estimation. While such systems can improve downstream spatial judgments, they do not address whether perspective-taking can arise within a model's internal representations. In contrast, our intervention modifies only the tokenization and training curriculum, allowing allocentric reasoning to emerge endogenously within the model. This distinction has broader implications for interpretability and cognitive alignment. Perspective-taking becomes a property of the model's own representational geometry rather than an artifact of post-hoc processing. Critically, perspective tokens illuminate what a model must represent to perform VPT, not just how to succeed on a specific benchmark.

Although our experiments focus on LLaVA 1.5 13B, the token-based mechanism we introduce is model-agnostic, requires no architectural modifications, and integrates naturally into any transformer-based multimodal model with a trainable tokenizer. This suggests that structured spatial tokenization may provide a generalizable strategy to enable allocentric reasoning between model families. At the same time, our rotation tokens operate over two-dimensional spatial relations, whereas natural visual environments require reasoning about depth and three-dimensional structure. A promising direction for future work involves integrating perspective tokens with depth-based perception tokens to create comprehensive three-dimensional spatial representations. By augmenting object representations with depth information—effectively adding a third spatial dimension along with planar position and orientation—models could reason about object locations relative to different viewpoints in full three-dimensional space. This would enable more sophisticated spatial reasoning tasks, such as understanding occlusion from different perspectives, judging relative distances between objects from another's viewpoint, or navigating complex three-dimensional environments. Recent work has shown that perception tokens can encode fine-grained depth information \citep{bigverdi2024perceptiontokensenhancevisual}, and integrating such representations with our orientation-based tokens offers a principled path toward unified three-dimensional allocentric reasoning. More broadly, this approach aligns with human spatial cognition, in which depth perception and perspective-taking operate jointly to support navigation and social interaction in three-dimensional space.

Together, these results illustrate how insights from cognitive psychology can inform design and analysis of multimodal language models, while computational models provide a testbed for evaluating theories of spatial reasoning. By grounding our intervention in established cognitive mechanisms, we show how imposing cognitively motivated structure on internal representations can selectively enable complex forms of spatial reasoning. This work exemplifies a bidirectional exchange between cognitive science and machine intelligence, in which theories of human cognition guide representational design. More broadly, such mechanistic alignment offers a path toward building multimodal systems whose internal reasoning more closely reflects human strategies.

\section*{Methods}

\subsection*{Token Construction}
All training images came from a subset of COCO 2017 train images \citep{lin2015microsoftcococommonobjects} containing only one person in the scene identified through the use of Voxel51 \citep{moore2020fiftyone}. Restricting images to single-reference scenes ensured unambiguous assignment of perspective during training. Before token creation, all images were resized to 336×336 pixels to place token coordinates within a fixed pixel range for training.

\subsubsection*{Embodiment Tokens}
Embodiment tokens rely on body keypoints of the reference person in the image to identify their angle of orientation. We generated two sets of pose token data using keypoints from different sources: (1) keypoints available directly in the COCO dataset and (2) keypoints generated by ViT Pose \citep{xu2022vitpose}, allowing us to evaluate whether embodiment tokens remain effective when ground-truth annotations are unavailable. ViTPose produced highly similar keypoint predictions to COCO, with an average difference of 7.2 pixels ($\approx 2\%$ of image width). This process is visually described in Figure \ref{fig1}B. 

Embodiment tokens prompt the model to first identify main body keypoints (right and left shoulder and hip joints) of the reference to represent the orientation of the reference’s torso and estimate their yaw, or rotation around the vertical axis. To calculate pose tokens, we first calculate the pixel difference between the right and left shoulder coordinates as
\[
\Delta x = x_R - x_L, \quad \Delta y = y_R - y_L .
\]  
Because image coordinates are defined with the \(y\)-axis increasing downward, the sign of \(\Delta y\) was inverted prior to computing orientation. The yaw angle of the torso is then computed as  
\[
\theta = \big( \arctan2(-\Delta y, \, \Delta x) \cdot \tfrac{180}{\pi} + 360 \big) \bmod 360 .
\]  
Finally, to obtain a discrete representation, we divide the circle into \(B\) equal bins (here, \(B=8\)), and assign each sample to a yaw token  
\[
\text{YAW}_k, \quad k = \left\lfloor \frac{\theta}{360 / B} \right\rfloor .
\]  

During token generation, the model is supervised to first predict the body keypoint tokens (e.g., left and right shoulder locations) before predicting the corresponding yaw token. As a result, the model can learn an explicit geometric relationship between keypoint configuration and orientation: when the right shoulder appears at a larger x-coordinate than the left shoulder ($\Delta x > 0$), the torso is facing away from the viewer, corresponding to aligned yaw bins (e.g., $k \in \{0,1,7\}$. Conversely, when the right shoulder appears to the left of the left shoulder ($\Delta x < 0$), the torso is towards the viewer, corresponding to unaligned yaw bins. This supervision encourages the model to internalize a simple, physically grounded rule linking relative keypoint geometry to egocentric alignment, rather than treating orientation as an unstructured categorical label.

This discretization allowed orientation information to be represented as categorical tokens while preserving sensitivity to coarse angular differences. References whose left keypoints appear to the left of their right keypoints in the image (indicating they face the same direction as the viewer) would be considered aligned, whereas references whose left keypoints appear to the right of their right keypoints (indicating they face away) would be considered unaligned.

\subsubsection*{Rotation Tokens}

Rotation tokens encode spatial information more abstractly by directly labeling both the reference and query objects with an (x, y) coordinate and an azimuth value corresponding to the object's facing direction. Unlike embodiment tokens, rotation tokens do not rely on body-based representations and can therefore be applied to non-human references. Our approach follows the orientation encoding method introduced by \cite{lee2025perspectiveawarereasoningvisionlanguagemodels}, where (x, y) coordinates were derived from the mean of each object's bounding box and azimuths were extracted from cropped object images using OrientAnything \citep{orient_anything}. This process is visually described in Fig \ref{fig1}b.

\subsection*{Training Protocol}

We fine-tuned the language model component using LoRA \citep{hu2021loralowrankadaptationlarge} to enable parameter-efficient learning. During training, the vision backbone was kept fixed, serving as a frozen feature extractor, while the language model was adapted via LoRA modules. In addition, the token embedding matrix and the language modeling head were fully trainable to support the introduction of new spatial tokens. This setup allowed gradients to propagate through the full model while restricting parameter updates to the language model adapters and token-level parameters.

Inspired by \cite{bigverdi2024perceptiontokensenhancevisual}, we followed a curriculum learning approach for structuring our training data. This approach was designed to separate learning how to generate spatial tokens from learning how to reason with them. Models were first trained to generate perception tokens directly from images, and then progressed to multi-step reasoning where they integrated these tokens to solve perspective-taking tasks.
For embodiment tokens, we created 18,000 token generation examples, with 200 chain-of-thought and 200 direct answer examples drawn from this set. For rotation tokens, we created 20,000 token generation examples, with 650 chain-of-thought and 650 direct answer examples. We implemented an annealing schedule to gradually transition from token generation to reasoning: across 10 epochs, the proportion of token generation examples decreased linearly from 100\% to 10\% (in 10\% decrements per epoch), while chain-of-thought and direct answer examples increased correspondingly. Chain-of-thought examples required the model to explicitly reason through perspective transformations using the tokens, whereas direct-answer examples prompted the model to produce spatial judgments (e.g., "left" or "right") without intermediate reasoning steps (Fig. \ref{fig1}c). This gradual curriculum enabled the model to first establish robust token representations before learning to apply them in reasoning contexts.

To support specialized spatial representations, we expanded LLaVA's tokenizer to include new spatial tokens. COCO embodiment tokens consisted of 336 tokens each for x and y coordinates (672 total), 8 yaw tokens, 4 torso width tokens, 4 tokens denoting the start and end of pose/orientation sequences, and 4 tokens for body keypoints (total: 692 tokens). ViTPose embodiment tokens included the same components plus 10 additional tokens representing confidence ratings for each keypoint (total: 702 tokens). Rotation tokens consisted of 672 x-y coordinate tokens, 18 semantic category tokens for object groups (e.g., person, furniture), 10 azimuth tokens representing 36° orientation bins, and 2 tokens for object sequence boundaries (total: 702 tokens).

\subsection*{Evaluation Protocol}

\subsubsection*{Perspective-Taking Benchmark}
Our primary evaluation used a perspective-taking benchmark designed to assess level 2 VPT in both humans and AI systems \citep{leonard2024failuresperspectivetakingmultimodalai}. Unlike prior benchmarks, this task systematically controls reference orientation and object placement using methods adapted from human psychophysics. Stimuli were created using Blender and varied the reference avatar's orientation across angles while balancing the spatial location of target objects (a sphere and cube) to control for visual biases. This design allows precise identification of egocentric errors and their correction by token-based interventions.

\subsubsection*{Isle Bricks V2}
\cite{góral2025recognitionevaluatingvisualperspective} introduced Isle Bricks V2 as an extension of their first Isle dataset \citep{góral2024seeingeyesevaluatingvisual}. Isle Bricks contains images of scenes with humanoid minifigures and single objects varying both object position and the minifigure's orientation. Filtering these items for queries involving left/right spatial judgments resulted in 40 evaluation items.

\subsubsection*{COCO 2017 validation images}
To evaluate on more naturalistic images, we created a small evaluation dataset using single-person images from the COCO 2017 validation data (training data were made from the training set). After filtering for items with correct orientation from OrientAnything, we were left with 45 evaluation items.

\subsubsection*{3DSRBench}
\cite{ma20253dsrbenchcomprehensive3dspatial} introduced 3DSRBench as a comprehensive spatial reasoning benchmark for multimodal models. The benchmark includes four types of spatial reasoning questions, including an orientation subset (Fig. \ref{fig1}d) that involves perspective-taking on naturalistic COCO images. We selected all items in this category and filtered them for queries that involved relevant references like people, animals, and chair-like furniture, yielding 45 evaluation items.

\subsubsection*{Feature Analysis}
To investigate how internal representations change with fine-tuning, we analyzed hidden-unit activations in response to controlled stimulus features from the perspective-taking benchmark (avatar alignment, angular difference, and cube direction; Fig. \ref{fig3}a). Hidden states were extracted using PyTorch forward hooks \citep{NEURIPS2019_bdbca288} from all perspective-taking stimuli. For each hooked module, activations of shape \textit{(batch\_size, seq\_length, hidden\_dim)} were averaged across the sequence dimension, yielding a two-dimensional representation. We performed this analysis on two layers: the multimodal projector output (model.mm\_projector[2]) and the final language-model attention projection (model.layers[-1].self\_attn.q\_proj). Examining both layers allowed us to assess how spatial information is represented at the multimodal interface and within language-only processing stages.

To facilitate comparisons across units with differing activation scales, unit activations were standardized (z-scored) across all stimuli prior to statistical analysis.

To identify units selectively sensitive to task-relevant features, we tested each hidden unit for statistically significant modulation across experimental conditions (e.g., aligned vs. unaligned reference orientations). Specifically, we compared activation magnitudes using Welch's t-test with a significance threshold of p $<$ 0.05. Units showing significantly greater activation in a given contrast were classified as feature-specific units. The number of feature-specific units served as our primary measure of how extensively task-relevant features are encoded within each model's representations. This analysis parallels feature-selectivity approaches in neuroscience while avoiding assumptions about spatial contiguity or modular organization within transformer architectures.

\section*{Data Availability}

The perspective-taking dataset referenced in this study is publicly available via HuggingFace at \\ \href{https://huggingface.co/datasets/bridgetleonard/PerspectiveTaking}{https://huggingface.co/datasets/bridgetleonard/PerspectiveTaking}. This study also uses publicly available datasets and evaluation benchmarks, including the Isle Dataset (\href{https://huggingface.co/datasets/Gracjan/Isle}{https://huggingface.co/datasets/Gracjan/Isle}), 3DSRBench (\href{https://huggingface.co/datasets/ccvl/3DSRBench}{https://huggingface.co/datasets/ccvl/3DSRBench}), and COCO (\href{https://cocodataset.org}{https://cocodataset.org}), which are available at their respective repositories.  

\section*{Code Availability}

The full code for reproducing our results is available at \href{https://github.com/bridgetleonard2/EmbodimentTokens}{https://github.com/bridgetleonard2/EmbodimentTokens} for the embodiment token pipeline and \href{https://github.com/bridgetleonard2/RotationTokens}{https://github.com/bridgetleonard2/RotationTokens} for rotation tokens.

\section*{Acknowledgments}
We thank K. Woodard for creating the perspective-taking stimuli in Blender; R. Krishna and M. Bigverdi for sharing code for fine-tuning LLaVA with new tokens and for helpful discussions; and G. Góral and W. Ma for developing the Isle dataset and 3DSRBench, respectively. This work was supported by NIH grant R01MH131595 (S.O.M.).

\bibliography{references}

@article{GUNIA2021113247,
title = {Brain mechanisms of visuospatial perspective-taking in relation to object mental rotation and the theory of mind},
journal = {Behavioural Brain Research},
volume = {407},
pages = {113247},
year = {2021},
issn = {0166-4328},
doi = {https://doi.org/10.1016/j.bbr.2021.113247},
url = {https://www.sciencedirect.com/science/article/pii/S0166432821001352},
author = {Anna Gunia and Sofiia Moraresku and Kamil Vlček},
}

@inproceedings{NEURIPS2019_bdbca288,
 author = {Paszke, Adam and Gross, Sam and Massa, Francisco and Lerer, Adam and Bradbury, James and Chanan, Gregory and Killeen, Trevor and Lin, Zeming and Gimelshein, Natalia and Antiga, Luca and Desmaison, Alban and Kopf, Andreas and Yang, Edward and DeVito, Zachary and Raison, Martin and Tejani, Alykhan and Chilamkurthy, Sasank and Steiner, Benoit and Fang, Lu and Bai, Junjie and Chintala, Soumith},
 booktitle = {Advances in Neural Information Processing Systems},
 editor = {H. Wallach and H. Larochelle and A. Beygelzimer and F. d\textquotesingle Alch\'{e}-Buc and E. Fox and R. Garnett},
 pages = {},
 publisher = {Curran Associates, Inc.},
 title = {PyTorch: An Imperative Style, High-Performance Deep Learning Library},
 url = {https://proceedings.neurips.cc/paper_files/paper/2019/file/bdbca288fee7f92f2bfa9f7012727740-Paper.pdf},
 volume = {32},
 year = {2019}
}

@article{liu2023improvedllava,
      title={Improved Baselines with Visual Instruction Tuning}, 
      author={Liu, Haotian and Li, Chunyuan and Li, Yuheng and Lee, Yong Jae},
      journal={Preprint at \url{https://arxiv.org/abs/2310.03744}},
      year={2023},
}

@article{zhang2025sphereunveilingspatialblind,
      title={SPHERE: Unveiling Spatial Blind Spots in Vision-Language Models Through Hierarchical Evaluation}, 
      author={Wenyu Zhang and Wei En Ng and Lixin Ma and Yuwen Wang and Junqi Zhao and Allison Koenecke and Boyang Li and Lu Wang},
      journal={Preprint at \url{https://arxiv.org/abs/2412.12693}},
      year={2025}
}

@article{zhang2025visionlanguagemodelsrepresentspace,
      title={Do Vision-Language Models Represent Space and How? Evaluating Spatial Frame of Reference Under Ambiguities}, 
      author={Zheyuan Zhang and Fengyuan Hu and Jayjun Lee and Freda Shi and Parisa Kordjamshidi and Joyce Chai and Ziqiao Ma},
      journal={Preprint at \url{https://arxiv.org/abs/2410.17385}},
      year={2025},
}

@article{leonard2024failuresperspectivetakingmultimodalai,
      title={Failures in Perspective-taking of Multimodal AI Systems}, 
      author={Bridget Leonard and Kristin Woodard and Scott O. Murray},
      journal={Preprint at \url{https://arxiv.org/abs/2409.13929}},
      year={2024},
}

@article{góral2024seeingeyesevaluatingvisual,
      title={Seeing Through Their Eyes: Evaluating Visual Perspective Taking in Vision Language Models}, 
      author={Gracjan Góral and Alicja Ziarko and Michal Nauman and Maciej Wołczyk},
      journal={Preprint at \url{https://arxiv.org/abs/2409.12969}},
      year={2024},
}

@article{góral2025recognitionevaluatingvisualperspective,
      title={Beyond Recognition: Evaluating Visual Perspective Taking in Vision Language Models}, 
      author={Gracjan Góral and Alicja Ziarko and Piotr Miłoś and Michał Nauman and Maciej Wołczyk and Michał Kosiński},
      journal={Preprint at \url{https://arxiv.org/abs/2505.0382}},
      year={2025},
}

@article{lee2025perspectiveawarereasoningvisionlanguagemodels,
      title={Perspective-Aware Reasoning in Vision-Language Models via Mental Imagery Simulation}, 
      author={Phillip Y. Lee and Jihyeon Je and Chanho Park and Mikaela Angelina Uy and Leonidas Guibas and Minhyuk Sung},
      journal={Preprint at \url{https://arxiv.org/abs/2504.17207}},
      year={2025},
}

@article{bigverdi2024perceptiontokensenhancevisual,
      title={Perception Tokens Enhance Visual Reasoning in Multimodal Language Models},
      author={Bigverdi, Mahtab and Luo, Zelun and Hsieh, Cheng-Yu and Shen, Ethan and Chen, Dongping and Shapiro, Linda G and Krishna, Ranjay},
      journal={Preprint at \url{https://arxiv.org/abs/2412.03548}},
      year={2024},
}

@article{ma20253dsrbenchcomprehensive3dspatial,
      title={3DSRBench: A Comprehensive 3D Spatial Reasoning Benchmark}, 
      author={Wufei Ma and Haoyu Chen and Guofeng Zhang and Yu-Cheng Chou and Jieneng Chen and Celso M de Melo and Alan Yuille},
      journal={Preprint at \url{https://arxiv.org/abs/2412.07825}},
      year={2025},
}

@article{moll2006level,
  title={Level 1 perspective-taking at 24 months of age},
  author={Moll, Henrike and Tomasello, Michael},
  journal={British Journal of Developmental Psychology},
  volume={24},
  number={3},
  pages={603--613},
  year={2006},
  publisher={Wiley Online Library}
}

@article{kessler2010embodied,
  title={The embodied nature of spatial perspective taking: Embodied transformation versus sensorimotor interference},
  author={Kessler, Klaus and Thomson, Lindsey Anne},
  journal={Cognition},
  volume={114},
  number={1},
  pages={72--88},
  year={2010},
  publisher={Elsevier}
}

@article{newcombe1992children,
  title={Children's early ability to solve perspective-taking problems.},
  author={Newcombe, Nora and Huttenlocher, Janellen},
  journal={Developmental psychology},
  volume={28},
  number={4},
  pages={635},
  year={1992},
  publisher={American Psychological Association}
}

@article{surtees2012egocentrism,
  title={Egocentrism and automatic perspective taking in children and adults},
  author={Surtees, Andrew and Apperly, Ian},
  journal={Child development},
  volume={83},
  number={2},
  pages={452--460},
  year={2012},
  publisher={Wiley Online Library}
}

@article{dumontheil2010taking,
  title={Taking perspective into account in a communicative task},
  author={Dumontheil, Iroise and K{\"u}ster, Olivia and Apperly, Ian A and Blakemore, Sarah-Jayne},
  journal={Neuroimage},
  volume={52},
  number={4},
  pages={1574--1583},
  year={2010},
  publisher={Elsevier}
}

@article{surtees2013similarities,
  title={Similarities and differences in visual and spatial perspective-taking processes},
  author={Surtees, Andrew and Apperly, Ian and Samson, Dana},
  journal={Cognition},
  volume={129},
  number={2},
  pages={426--438},
  year={2013},
  publisher={Elsevier}
}

@article{surtees2013use,
  title={The use of embodied self-rotation for visual and spatial perspective-taking},
  author={Surtees, Andrew and Apperly, Ian and Samson, Dana},
  journal={Frontiers in human neuroscience},
  volume={7},
  pages={698},
  year={2013},
  publisher={Frontiers Media SA}
}

@article{lin2015microsoftcococommonobjects,
      title={Microsoft COCO: Common Objects in Context}, 
      author={Tsung-Yi Lin and Michael Maire and Serge Belongie and Lubomir Bourdev and Ross Girshick and James Hays and Pietro Perona and Deva Ramanan and C. Lawrence Zitnick and Piotr Dollár},
      journal={Preprint at \url{https://arxiv.org/abs/1405.0312}},
      year={2015},
}

@article{moore2020fiftyone,
  title={FiftyOne},
  author={Moore, B. E. and Corso, J. J.},
  journal={GitHub. Note: https://github.com/voxel51/fiftyone},
  year={2020}
}

@inproceedings{
  xu2022vitpose,
  title={Vi{TP}ose: Simple Vision Transformer Baselines for Human Pose Estimation},
  author={Yufei Xu and Jing Zhang and Qiming Zhang and Dacheng Tao},
  booktitle={Advances in Neural Information Processing Systems},
  year={2022},
}

@article{orient_anything,
  title={Orient Anything: Learning Robust Object Orientation Estimation from Rendering 3D Models},
  author={Wang, Zehan and Zhang, Ziang and Pang, Tianyu and Du, Chao and Zhao, Hengshuang and Zhao, Zhou},
  journal={Preprint at \url{https://arXiv.org/abs/2412.18605}},
  year={2024}
}

@article{hu2021loralowrankadaptationlarge,
      title={LoRA: Low-Rank Adaptation of Large Language Models}, 
      author={Edward J. Hu and Yelong Shen and Phillip Wallis and Zeyuan Allen-Zhu and Yuanzhi Li and Shean Wang and Lu Wang and Weizhu Chen},
      journal={Preprint at \url{https://arxiv.org/abs/2106.09685}},
      year={2021},
}

\end{document}